\pdfoutput=1

\documentclass[11pt]{article}

\usepackage[]{acl}

\usepackage{times}
\usepackage{latexsym}
\usepackage{amsmath}
\usepackage{booktabs}
\usepackage{graphicx}
\usepackage{multirow}
\usepackage{bm}

\usepackage{algorithm}
\usepackage{algorithmic}
\usepackage{subfigure}
\usepackage{physics}
\newlength{\lsuperstar}
\usepackage[T1]{fontenc}

\usepackage[utf8]{inputenc}

\usepackage{microtype}

\title{Model Uncertainty--Aware Knowledge Amalgamation for \\Pre-Trained Language Models}

\author{Lei Li\textsuperscript{$\dag$}, Yankai Lin\textsuperscript{$\S$}, Xuancheng Ren\textsuperscript{$\dag$}, \\
\textbf{Guangxiang Zhao\textsuperscript{$\dag$}, Peng Li\textsuperscript{$\S$}, Jie Zhou\textsuperscript{$\S$}, Xu Sun\textsuperscript{$\dag$}} \\
   \textsuperscript{$\dag$}MOE Key Laboratory of Computational Linguistics, School of EECS, Peking University \\
  \textsuperscript{$\S$}Pattern Recognition Center, WeChat AI, Tencent Inc., China\\
    \texttt{lilei@stu.pku.edu.cn}\\  \texttt{\{renxc, guangxiangzhao, xusun\}@pku.edu.cn} \\
    \texttt{\{yankailin, patrickpli, withtomzhou\}@tecent.com}
  }

\begin{document}
\maketitle
\begin{abstract}
As many fine-tuned pre-trained language models~(PLMs) with promising performance are generously released, investigating better ways to reuse these models is vital as it can greatly reduce the retraining computational cost and the potential environmental side-effects.
In this paper, we explore a novel model reuse paradigm, Knowledge Amalgamation~(KA) for PLMs. Without human annotations available, KA aims to merge the knowledge from different teacher-PLMs, each of which specializes in a different classification problem, into a versatile student model.
The achieve this, we design a Model Uncertainty--aware Knowledge Amalgamation~(MUKA) framework, which identifies the potential adequate teacher using Monte-Carlo Dropout
for approximating the golden supervision to guide the student.
Experimental results demonstrate that MUKA achieves substantial improvements over baselines on benchmark datasets.
Further analysis shows that MUKA can generalize well under several complicate settings with multiple teacher models, heterogeneous teachers, and even cross-dataset teachers.
\end{abstract}

\section{Introduction}

Large-scale pre-trained language models~(PLMs), such as BERT~\citep{devlin2019bert}, RoBERTa~\citep{Liu2019RoBERTa} and T5~\citep{raffel20t5} have recently achieved promising results after fine-tuning on various natural language processing~(NLP) tasks.
Many fine-tuned PLMs are generously released for facilitating researches and deployments.
Reusing these PLMs can greatly reduce the computational cost of retraining the PLM from scratch, thus alleviating the potential environmental side-effects~\citep{strubell-etal-2019-energy,schwartz2020green}.

\begin{figure}[t!]
    \centering
    \includegraphics[width=0.95\linewidth]{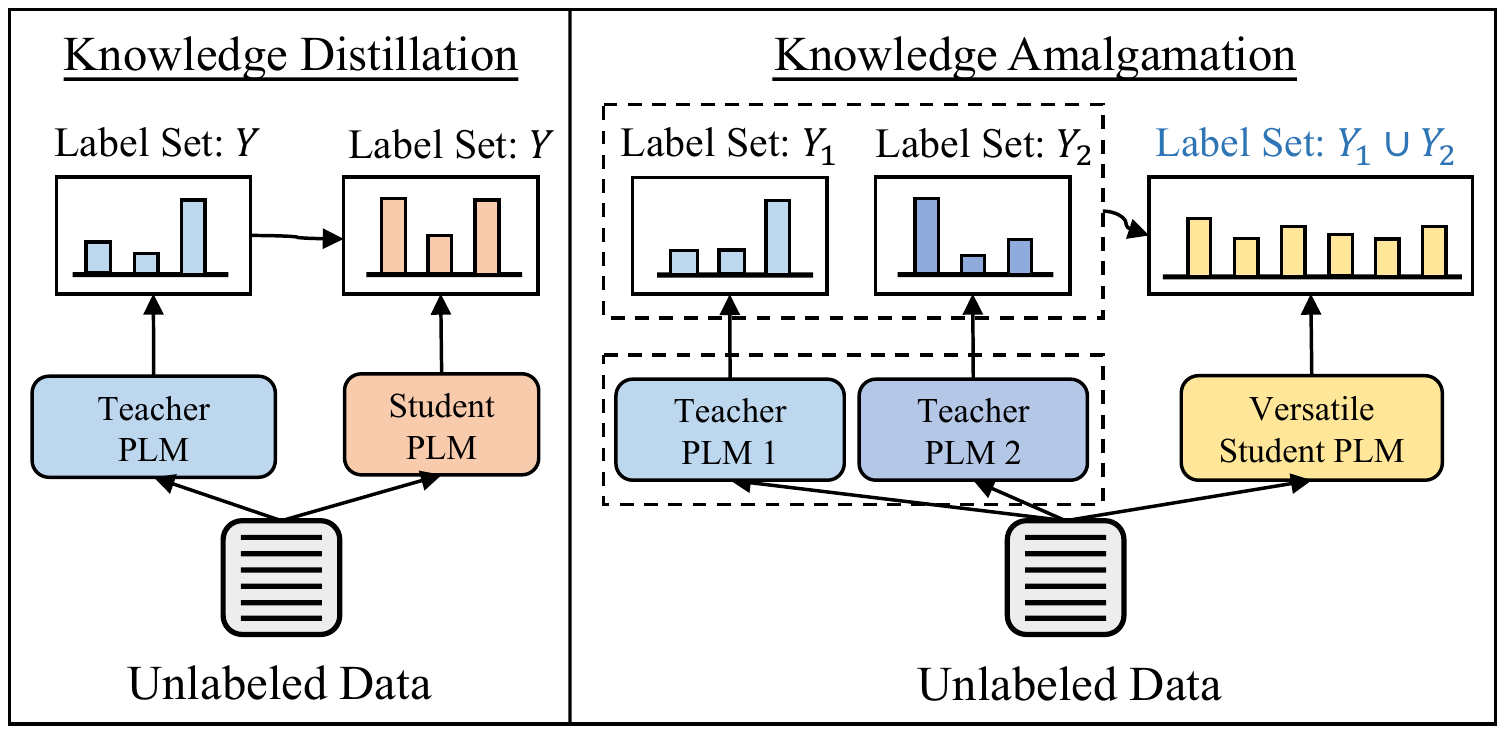}
    \caption{Comparison of knowledge distillation~(KD) and knowledge amalgamation~(KA) with two teachers. KD assumes that the student performs predictions on the identical label set with the teacher, while KA trains a student model that is capable of performing classification over the union label set of teacher models.}
    \label{fig:ka_tc}
\end{figure}

A commonly adopted model reuse paradigm is knowledge distillation~\citep{Hinton2015Distilling,romero15fitnet}, which transfers learned knowledge from a teacher model to a student model via matching the outputs between the teacher model and the student. 
Though achieving promising results with PLMs~\citep{Sun2019PatientKD,Jiao2019TinyBERT,Sanh2019DistilBERT}, KD restricts the student to perform the same task as the teacher model, leading to an under-utilization of released PLMs fine-tuned on different tasks.
In this paper, we explore a novel model reuse paradigm for PLMs, namely \emph{Knowledge Amalgamation}~(KA). 
Given multiple fine-tuned teacher-PLMs, each of which is capable of performing classification over a unique label set, KA aims to train a versatile student that can make predictions over the union of teacher label sets.
Note that we assume no human annotations are available during KA, as in many scenarios the labeled data for training the teachers may not be publicly available due to privacy issues.
KA can make full use of the released PLMs, as it relaxes the assumption of KD and trains a student to cover the classification skills of all teacher models.
Figure~\ref{fig:ka_tc} illustrates the difference between KA and KD.



As no annotations are available, the core challenge of KA is to construct supervision to guide the student.
To this end, we propose a principled framework to achieve KA for PLMs.
Assuming that there is an oracle teacher capable of providing virtual supervision over the union label set, through theoretical derivation, we find that the key to construct such supervision is to identify the adequate teacher for each instance.
Our investigation shows that model uncertainty estimated with Monte-Carlo Dropout~\citep{gal2016dropout} can be utilized as a good proxy.
We then build our Model Uncertainty--aware Knowledge Amalgamation~(MUKA) framework, where the golden supervision is approximated by either taking the outputs of the most confident teacher, or softly integrating different teacher predictions according to the relative importance of each teacher.
Furthermore, an instance re-weighting mechanism based on the margin of uncertainty scores is introduced, to down-weight the contribution of instances with confusing supervision signals.
Experimental results show that MUKA can significantly outperform previous baselines, even achieving  comparable  results with models trained with labeled data.
Further analysis shows that MUKA can produce supervision close to the golden distribution and generalize well in challenging settings, including merging knowledge from multiple teacher models, heterogeneous teachers with different architectures, or cross-dataset even cross-lingual teacher models. 

\begin{figure*}[t!]
    \centering
    \includegraphics[width=0.9\linewidth]{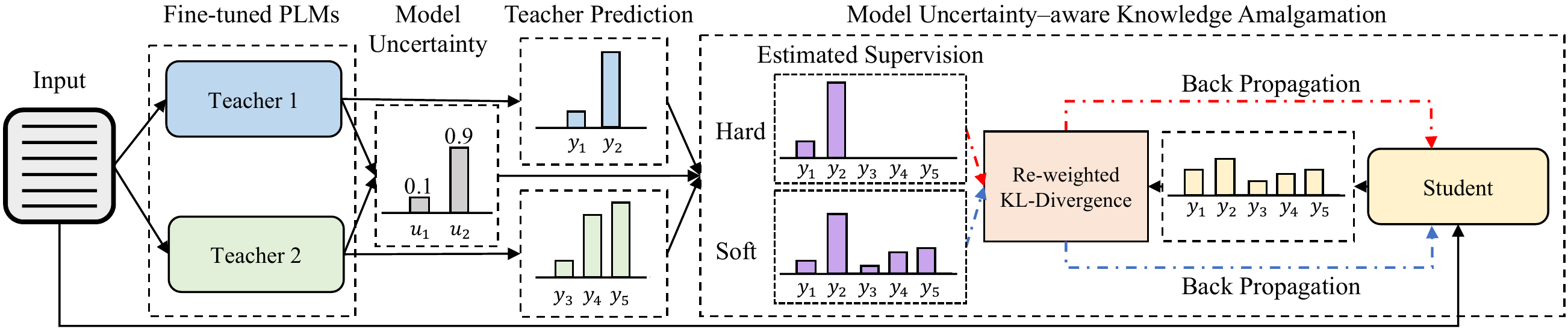}
    \caption{Overview of the proposed MUKA framework. The label distribution over the union label set is estimated according to the model uncertainty scores. 
    The student model is trained via minimizing the KL-divergence between the hard(red) /soft(blue) estimated supervision and the student output, weighted by the model uncertainty margin of teachers. Best viewed in color.
    }
    \label{fig:uka_framework}
\end{figure*}

The main contributions of this work can be
summarized as follows:
(1) We explore a novel model reuse paradigm, knowledge amalgamation for PLMs, making full use of released PLMs with different label sets.
(2) We present MUKA, which approximates the golden supervision according to model uncertainty and re-weights the instance contribution based on uncertainty margin.
(3) Extensive results on benchmarks demonstrate that MUKA  provides more accurate supervision, and is generalizable for amalgamating heterogeneous teachers and cross-dataset teachers.

\section{Related Work and Background}
\paragraph{Knowledge Distillation~(KD)} transfers the knowledge from a teacher model to a student model.
For example, \citet{Hinton2015Distilling} utilize the soft labels produced by the teacher model for the student to learn.
and \citet{romero15fitnet} align the intermediate representations between the student and the teacher.
Recent studies apply KD for PLMs by matching the intermediate states~\citep{Sun2019PatientKD,Sanh2019DistilBERT,wang2020minilm}, enriching the training data~\citep{Jiao2019TinyBERT,liang2020mixkd}, and learning from multiple teachers~\citep{wu-etal-2021-one,li-etal-2021-dynamic}. 
However, all these KD studies assume that the student has an identical label set with the target teacher model(s), restricting better utilization of PLMs with different label sets.

\paragraph{Knowledge Amalgamation~(KA)} aims to train a versatile student model with unlabeled data, by amalgamating the knowledge from multiple teachers. 
Previous studies mainly focus on reusing teacher CNNs for image-related tasks, and can be roughly categorized into two streams.
One is matching the student outputs separately to the corresponding teacher~\citep{Vongkulbhisal19UHC}.
The other instead focuses on aligning the representations of the student and the teachers:
\citet{shen19comprehensive} map the student features to that of the teacher model layer-by-layer and \citet{Luo19CFL} conduct the feature alignment in a common space by minimizing the maximum mean discrepancy.
However, the former results in conflicting supervision for the student as teacher-PLMs tend to be overconfident~\citep{desai2020calibration}, and the latter paradigm requires complicated optimization design for generalizing to heterogeneous teachers.
To the best of our knowledge, we are the first to explore KA for PLMs. We propose a principled framework based on model uncertainty and demonstrate its effectiveness and generalizability on benchmark datasets.
\paragraph{Bayesian Neural Networks~(BNN)} replaces a deterministic weight parameter of a model with a prior distribution of the weight.
Instead of optimizing the model weights directly, BNN averages over all possible weights when conducting inference.
Formally, Bayesian inference aims to estimate the posterior distribution $p\left(\mathbf{W} \mid \mathcal{D}\right)$ of the model parameters $\mathbf{W}$ given the dataset $\mathcal{D}$.
The prediction of Bayesian inference for input $x$ is conducted by marginalizing over all the possible parameters as:
\begin{equation}
p\left(y \mid x, \mathcal{D}\right)=\int p\left(y \mid \mathbf{\mathbf{W}}, x\right) p(\mathbf{\mathbf{W}} \mid \mathcal{D}) \dd \mathbf{W} .
\end{equation}
As the exact integration is intractable for deep neural networks with complex architectures and high dimension parameters,
various variational inference methods~\citep{graves2011practical,blundell2015weight,gal2016dropout} have been proposed to find a surrogate distribution $q_\theta\left( \mathbf{W}\right)$ to approximate the underlying posterior $p\left(\mathbf{W} \mid \mathcal{D}\right)$, where the Kullback-Leibler~(KL) divergence of these two distributions are minimized. 
Among these, the Monte-Carlo dropout method~\citep{gal2016dropout} is widely adopted as it requires minimal changes to the original model, which takes the Dropout distribution~\citep{srivastava2014dropout} as $q_\theta({\mathbf{W}})$ to sample $K$ masked model weights $\left\{ \mathbf{W}_k\right\}_{k=1}^{K} \sim q_\theta\left(\mathbf{W}\right)$.
The estimated probability for classification tasks thus can be calculated as:
\begin{align}
p\left(y \mid x, \mathcal{D}\right) & \approx \int p\left(y \mid \mathbf{\mathbf{W}}, x\right) q_\theta(\mathbf{W}) \dd\mathbf{W} \\ 
&\approx \frac{1}{K}\sum_{k=1}^{K} p\left(y \mid \mathbf{W}_{k}, x\right) .
\label{eq:mc}
\end{align}
\section{Methodology}
In this section, we first give the task formulation for knowledge amalgamation, followed by the elaboration on the proposed MUKA framework.
Figure~\ref{fig:uka_framework} gives an overview of our proposal. 

\subsection{Problem Formulation}
We explore the problem of knowledge amalgamation for text classification with PLMs.
Formally, given $N$ teacher PLM models $TS = \left\{T_1, \dots, T_N \right\}$, where each teacher $T_i$ specializes in a specific classification problem, i.e., a set of classes $Y_i$, a student model $S$ is supposed to be learned with an unlabeled dataset $\mathcal{D}$, which is able to perform predictions over the comprehensive class set $Y = \bigcup_{i=1}^{N} Y_{i}$.
As previous studies~\citep{Vongkulbhisal19UHC,Thadajarassiri21SKA} show that merging teachers with overlapping classes is easy, we thus focus on a more challenging yet practical setting where the teacher specialties are totally disjoint, i.e., $Y_i \cap Y_j = \emptyset, \forall i \neq j$.
\subsection{Model Uncertainty--Aware Knowledge Amalgamation}
As there are no annotated data available, we need to construct supervision for guiding the student.
Assuming that there is an oracle model $\mathcal{T}$, which is capable of giving a golden label distribution $\mathcal{T}(x)$ for each instance $x$ over $Y$, we can approximate this supervision, and then train the student by minimizing the KL-divergence between its prediction and the estimated distribution:

\begin{equation}
    \mathcal{L} = \sum_{x \in \mathcal{D}} \text{KL} \left(  S\left(x\right)  ||  \mathcal{T} \left(x\right)   \right),  
\label{eq:kd_loss}
\end{equation}
where $S(x)$ denotes the output distribution of the student for input $x$.
To estimate the golden supervision $\mathcal{T}(x)$, we first derive the correlation between $\mathcal{T}(x)$ and the prediction $T_i(x)$ of teacher model $T_i$.
Specifically, as teacher $T_i$ specializes in label set $Y_i$, it can only predict $T_i( y \mid x )$ for instance $x$ when $ y \notin Y_{-i} = \bigcup_{j=1, j \neq i}^{N} Y_j$.
Therefore, the correlation between $T_i( y  \mid x)$ and global probability $\mathcal{T}( y \mid x)$ over the full class set can be derived as: %
\begin{align}
T_{i}(y  \mid x ) 
&=\mathcal{T}\left(y  \mid x, y \in Y_{i}\right) \\
&=\frac{\mathcal{T}\left(y , y \in Y_{i} \mid x \right)}{\mathcal{T}\left( y \in Y_{i} \mid x \right)} . \label{eq:relation}
\end{align}
The above derivation indicates that we can recover the golden probability distribution by estimating the denominator, which means how likely the instance $x$ belongs to the classes in $Y_i$.
Followingly, we explore prediction uncertainty as an estimation of this probability.

\subsubsection{Uncertainty Estimation} %
\begin{figure}[t!]
    \centering
    \includegraphics[width=0.99\linewidth]{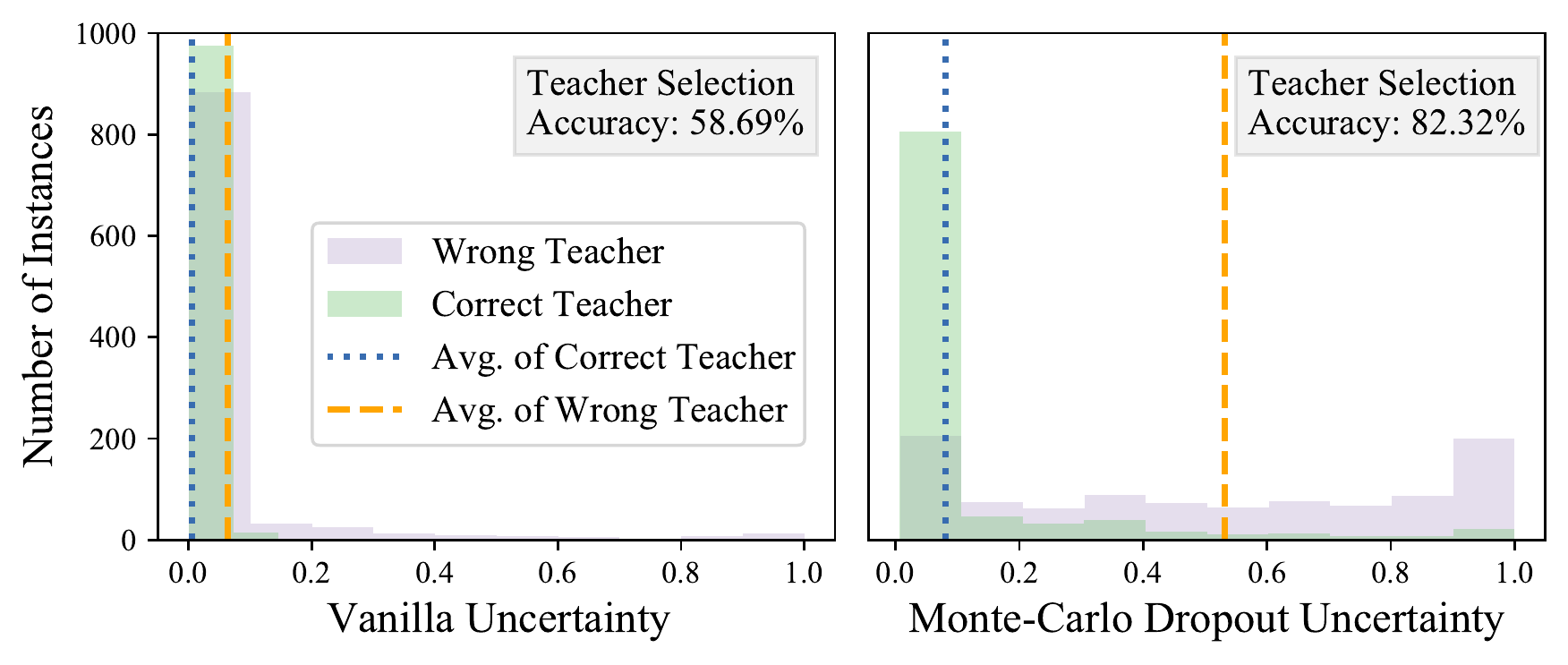}
    \caption{Model uncertainty~(normalized) distributions evaluated with $1000$ instances randomly sampled from the AG News dataset. The vanilla prediction entropy distributions of two teacher models overlaps greatly~(left), while Monte-Carlo Dropout produces a more accurate uncertainty approximation for distinguishing the correct teacher model~(right). Best viewed in color.
    }
    \label{fig:uncertainty_dist}
\end{figure}
As the instances associated with classes not in $Y_i$ can be treated as the out-of-distribution data for the teacher model $T_i$, the teacher predictions would be more confusing about these instances than that of in-distribution instances~\citep{hendrycks17baseline}.
Therefore, we propose to approximate the denominator in the Eq.~(\ref{eq:relation}) via model uncertainty.
A na\"ive estimation is taking the entropy of predicted class distribution.
However, due to the over-confident issues of over-parameterized models like PLMs~\citep{guo2017calibration,desai2020calibration}, this simple estimation can be unreliable.
We investigate this by first spliting the instances of the AG News dataset~\citep{cnn15zhang} into two sets with disjoint labels, and then fine-tuning teacher models on each set separately.
For each instance, there is a correct teacher that is capable of handling it and a wrong teacher that cannot make a correct prediction for it.
We plot the prediction entropy distributions of the correct teacher and the wrong teacher in the left part of Figure~\ref{fig:uncertainty_dist}.
It can be found that the wrong teacher also produces confident predictions for instances that are not in its speciality with nearly zero uncertainty scores, misleading the identification of the adequate teacher.

To remedy this, we propose to add small perturbations to the model parameter during prediction to find out the correct teacher model.
The intuition behind is that, as the instance is well fitted by the parameter of the correct teacher model, the adequate teacher thus produces confident results consistently in the multiple predictions even with small perturbed parameters. 
On the contrary, small perturbations on the model weights of the wrong teacher will lead to a drastic change in the output probabilities, resulting in more uncertainty predictions on average. 
Therefore, we can estimate the model uncertainty more accurately according to the average predictions under parameter perturbation, which is compatible with Monte-Carlo Dropout~\citep{gal2016dropout}, a Bayesian perspective for obtaining the model uncertainty score. 
Specifically, for an instance $x$, we calculate its output probability distribution over $Y_i$ with $T_i$ via Monte-Carlo Dropout as:
\begin{align}
p_i\left(y \mid x, \mathcal{D}\right)&\approx \frac{1}{K}\sum_{k=1}^{K} p_i\left(y \mid \mathbf{W}_{k}^i, x\right) \\
& = \frac{1}{K} \sum_{k=1}^K  T_i\left(x, \mathbf{W}_k^i \right),
\label{eq:mc_teacher}
\end{align}
where $\mathbf{W}_k^i$ is the $k$-th masked weights of $T_i$ sampled from the Dropout distribution~\citep{srivastava2014dropout}.
The model uncertainty of teacher model $T_i$ thus can be summarized as the entropy of the averaged probability distribution $p_i$:
\begin{equation}
    u_i = H \left( p_i\right) = - \sum_{y=1}^{ |Y_i|} p_i^y \log p_i^y.
    \end{equation}
As shown in the right part of Figure~\ref{fig:uncertainty_dist}, the uncertainty distribution estimated via Monte-Carlo Dropout exhibit a clear difference between the uncertainty distributions of the correct teacher and the wrong teacher model, indicating its great potential for guiding the probability combination.

With the accurately estimated teacher uncertainties $U = \{u_1, \dots, u_N \}$ at hand, we design two methods for approximating the golden supervision to guide the student model:

\noindent\textbf{MUKA-Hard} which directly takes the supervision provided by the teacher with the lowest uncertainty as the golden distribution:
\begin{equation}
\small 
    \mathcal{T}(x) = \text{Pad} \left( T_{i^*}\left(x\right)  \right) , 
    \text{where} \ i^* = \mathop{\arg\min}_i u_i/\log | Y_i | ,
\end{equation}
where $\log | Y_i | $ is a normalizing factor for comparison between teacher models with a different number of classes. 
As $T_i^*$ only provides the label relation over the class set $Y_{i^*}$, the probabilities of classes not in $Y_{i^*}$ are set to zeros, denoted by the $\text{Pad}$ operation.
In this way, we set $\mathcal{T}(y \in Y_{i^*} \mid x ) =1$, and thus the student can learn from the teacher model most confident about $x$.

\noindent\textbf{MUKA-Soft} which estimates the golden supervision as a weighted sum of teacher model predictions by taking the relative uncertain level into consideration:
\begin{align}
\small 
\mathcal{T}(x) &= \sum_{i=1}^{N}  w_i \text{Pad} \left(  T_i \left( x \right) \right) \\ 
w_i &= \frac{\exp\left(  c_i/ \tau\right)}{ \sum_{j=1}^N \exp(c_j/ \tau )} \label{eq:tau} \\
c_i &= 1 - u_i/\log |Y_i | ,
\end{align}
where $c_i$ denotes the confidence score which indicates how likely $x$ belongs to $C_i$, and $\tau$ is a hyper-parameter for controlling the smoothness of the weights.
In this way, the teacher with a higher confidence score contributes more to the estimated golden supervision signal.
Besides, the difference between the confidence scores reflects the inner correlation between the classes in different label groups, thus providing extra information for the classes in disjoint label sets. 

\subsubsection{Instance Re-weighting}
Furthermore, the uncertainty distribution overlapping in the right part of Figure~\ref{fig:uncertainty_dist} indicates that there is still a small portion of instances on which teacher models achieve similar confidence levels even with Monte-Carlo Dropout.
For these instances, MUKA-Hard may wrongly select the supervision source, and MUKA-Soft would assign close weights to all teachers, thus providing a vague supervision signal.
Therefore, the estimated supervision can be misleading for the student model.
To remedy this, we devise an instance re-weighting mechanism by modifying the objective in Eq.~(\ref{eq:kd_loss}):
\begin{align}
    \mathcal{L}_{\text{MUKA}} &=\sum_{x \in \mathcal{D}} v\left(x\right) \text{KL} \left(  S\left(x\right) || \mathcal{T}(x)\right) \\ 
    v\left(x\right) &=  c_{\text{max}} - c_{\text{sec}} ,
\end{align}
where $c_{\text{max}}$ and $c_{\text{sec}}$ denotes the largest and the second large teacher confidence score for instance $x$, respectively.
By minimizing the instance-level weighted objective, the student is encouraged to focus more on the pivotal instances with clearer supervision signals, thus achieving a better learning performance.

\section{Experiments}
\begin{table*}[t!]
    \centering
    \small 
    \begin{tabular}{@{}l| c | c c  c  c |c@{}}
    \toprule
       \textbf{Method}  &  \textbf{Model Size} & \textbf{AG News}  & \textbf{THUCNews} & \textbf{Google Snippets}& \textbf{5Abstracts Group}  & \textbf{Average}\\
       \midrule 
           Supervised & 110M  & 94.6 $\pm$ 0.00&  97.8 $\pm$ 0.00 & 89.3 $\pm$ 0.00&  90.7 $\pm$ 0.00 &  93.10 \\ 
      \midrule
     Teacher 1  & 110M   & 49.9 $\pm$ 0.00 &48.8 $\pm$ 0.00 & 50.2 $\pm$ 0.00 &42.0  $\pm$ 0.00 & 47.73\\ 
      Teacher 2 & 110M  & 47.5 $\pm$ 0.00 &49.8 $\pm$ 0.00 & 43.5 $\pm$ 0.00& 51.5 $\pm$ 0.00 & 48.08\\
      Ensemble &  220M & 59.8 $\pm$ 0.00& 93.1 $\pm$ 0.00  &80.4 $\pm$ 0.00  &  62.3 $\pm$ 0.00 &  73.90\\
      \midrule 
       Vanilla KD &  110M  &63.1 $\pm$  0.81 & 94.9 $\pm$ 0.18 & 83.7  $\pm$ 1.30 & 67.0 $\pm$ 1.14 &  77.18 \\
       DFA &  110M & 66.4 $\pm$ 2.33 & 94.4 $\pm$ 0.22 & 82.6 $\pm$ 0.18 & 57.7 $\pm$ 4.41 & 75.28\\ 
       CFL&  110M  & 61.4  $\pm$  1.18 & 95.1  $\pm$ 0.21& 84.5  $\pm$ 0.45&61.6  $\pm$ 0.12&  75.65 \\ 
       UHC &  110M  & 78.8 $\pm$  1.42& 92.1 $\pm$ 0.63&86.3 $\pm$ 0.39& 71.4 $\pm$ 0.67 &  82.15\\  
       \midrule 
       MUKA-Hard~(Ours) &  110M  &  87.0  $\pm$ 0.40 &  \textbf{97.2}  $\pm$ 0.12 & \textbf{88.4}  $\pm$  0.32&  79.0  $\pm$ 0.82 & \textbf{87.90}\\ 
        MUKA-Soft~(Ours) &  110M  & \textbf{87.1}  $\pm$ 0.19&  \textbf{97.2}  $\pm$ 0.08 & 87.9 $\pm$ 0.32 &  \textbf{79.3} $\pm$  0.85 & 87.88\\ 
    \bottomrule
    \end{tabular}
    \caption{Comparisons on the the benchmark datasets. 
    The results are classification accuracy averaged by three seeds, and standard deviations are reported. Both MUKA variants achieve statistically significant improvements over the best-performing baselines~($p < 0.01$). Best results are shown in bold. }
    \label{tab:main_ret}
\end{table*}
\subsection{Experimental Settings}
\paragraph{Datasets}
We conduct evaluations on four classic text classification benchmarks, including three English datasets: AG News~\citep{cnn15zhang}, Google Snippets~\citep{phan08gs}, 5Abstracts Group,\footnote{\url{https://github.com/qianliu0708/5AbstractsGroup}} and a Chinese dataset THUCNews~\citep{sun2016thuctc}. 
We split 5\% of data from the training set for datasets without a validation set to form a validation set for model selection.
The statistics of datasets can be found in Appendix~\ref{apx:dataset}.
\paragraph{Compared Methods}
We implement various baselines to evaluate the effectiveness of our proposal, which can be categorized as follows:

\emph{Simple Baselines}, which require no additional training, including:
(1) Original Teacher: The teacher models are used independently for prediction. We set the probabilities of classes out of the teacher speciality to zeros.
(2) Ensemble: The output logits of teachers are directly concatenated for predictions over the union label set.

\emph{KA Methods}, which assume internal states of the teacher model are available and the unlabeled data $\mathcal{D}$ are used to distill a student model, including:
(1) Vanilla KD~\citep{Hinton2015Distilling}: The student is trained to mimic the soft targets produced by logits combination of all teacher models, via minimizing the vanilla KL-divergence objective. 
(2) DFA~\citep{shen19comprehensive}: DFA designs a layer-wise feature adaptation mechanism for providing extra guidance based on Vanilla KD. The student aligns its features to the merged features of multiple teachers layer by layer.
(3) CFL~\citep{Luo19CFL}: CFL first maps the hidden representations of the student and the teachers into a common space. The student is trained by aligning the mapped features to that of the teachers, with supplemental supervision from the logits combination. 
(4) UHC~\citep{Vongkulbhisal19UHC}: UHC splits the student logits into subsets corresponding to the class sets of teacher models. Each subset is trained to mimic the corresponding output of the teacher model.

We also include a supervised learning method, which trains the student model with labeled data for better understanding the performance.

\paragraph{Implementation Details}
We implement our framework using the HuggingFace transformers library~\citep{wolf-etal-2020-transformers}.
The teacher and student models for English datasets and THUCNews are BERT-base-uncased~\citep{devlin2019bert} and BERT-wwm-ext~\citep{cui2020revisiting}, respectively.
For each dataset, the classes are randomly split into two non-overlapping parts, and two teachers are fine-tuned on each set separately to imitate the actual applications.
Detailed class split can be found in Appendix~\ref{apx:dataset}. 
We first fine-tune the teacher models with the split labeled data for $3$ epochs with a learning rate $2\times10^{-5}$. The trained teacher model weights are frozen during the student training process.
We set the forward number $K$ of Monte-Carlo Dropout uncertainty estimation to $16$ and the dropout rate is set to $0.1$.
Temperature $\tau$ in Eq.~(\ref{eq:tau}) is set to $0.2$ according to our hyper-parameter analysis results in Appendix~\ref{apx:temp}.
The student model then is learned by optimizing the KL-divergence objective for $3$ epochs, with a $2\times10^{-5}$ learning rate and $32$ batch size.
The student is evaluated on the validation set every $100$ step. We select the best performing checkpoints for final evaluation.
The experiments are replicated with $3$ random seeds and we report the averaged accuracy.

\subsection{Main Results}


The model performance comparison on the four datasets and the corresponding model size are listed in Table~\ref{tab:main_ret}.
Our findings are: 
(1) Simple baselines fall far behind, showing that it is necessary to conduct amalgamation. 
(2) DFA and CFL cannot achieve consistent improvements over Vanilla KD, demonstrating the instability of supervision based on feature alignments.
(3) UHC achieves better average results than Vanilla KD, while performs relatively worse on the THUCNews dataset, indicating its instability due to the potential supervision conflicting.
(4) Two variants of MUKA both outperform the previous baseline models on all the datasets, and the average accuracy of MUKA-Hard is achieves a $5.75$ points gain over the best performing baseline model. On the THUCNews dataset, while no label information is included during the knowledge amalgamation, MUKA can obtain a $97.2$ accuracy, which is very close to $97.8$ of the supervised learning method.
We attribute the success to that MUKA provides the student with the estimated golden probability distribution over the union label set according to model uncertainty, which can effectively transfer the knowledge and avoid potential supervision contradictory.
These promising results indicate that our MUKA framework can produce better supervisions for training the student model, thus has great potentials for reusing PLMs with different label sets.

\subsection{Ablation Studies}

\begin{table}[t!]
    \centering
    \small 
    \begin{tabular}{@{}l|c c@{}}
    \toprule 
    \textbf{Method}     &  \textbf{AG News}& \textbf{THUCNews} \\
    \midrule
    MUKA-Hard & 87.0 $\pm$  0.40 &  97.2 $\pm$ 0.12 \\ 
    \quad w/o Monte-Carlo Dropout& 65.1 $\pm$ 1.67 &97.0 $\pm$ 0.13\\ 
    \quad w/o Instance Re-weighting& 85.5  $\pm$ 0.51 &  96.9 $\pm$ 0.06  \\ 
    \midrule 
    MUKA-Soft & 87.1  $\pm$ 0.19 &  97.2  $\pm$  0.08 \\ 
    \quad w/o Monte-Carlo Dropout&  74.2  $\pm$ 0.28 &95.4  $\pm$ 0.20 \\ 
    \quad w/o Instance Re-weighting  &  86.7  $\pm$ 0.30 &  96.8  $\pm$ 0.03 \\ 
    \bottomrule
    \end{tabular}
    \caption{Ablation analysis of MUKA. The removed modules both lead to deteriorated performance.}
    \label{tab:ablation_study}
\end{table}

\paragraph{Contribution of Components in MUKA} We conduct ablation experiments on two large datasets for stable results, and the results are shown in Table~\ref{tab:ablation_study}.
By replacing the Monte-Carlo Dropout estimation of model estimation with a single forward estimation, i.e., set $K=1$ in Eq.~(\ref{eq:mc}), we find that the performance is degraded on both datasets.
This result validates our motivation to adopt the Monte-Carlo Dropout method to better estimate the model uncertainty.
Interestingly, we find that Monte-Carlo Dropout is much more effective on the AG News dataset.
To explore this, we compute the average ECE score~\citep{guo2017calibration} of two teacher models on the out-of-distribution samples, where higher ECE scores indicate more severe overconfident predictions.
The teacher models of AG News achieve an average $45.42$ ECE score, while that of THUCNews is $19.52$.
This result verifies the effectiveness of Monte-Carlo Dropout for accurate model uncertainty estimation of overconfident teachers.
We also find that removing the instance re-weighting mechanism leads to deteriorated results for both MUKA variants, demonstrating that paying more attention to instances with clearer supervision is beneficial for the student.

\begin{figure}[t]
    \centering
    \includegraphics[width=0.7\linewidth]{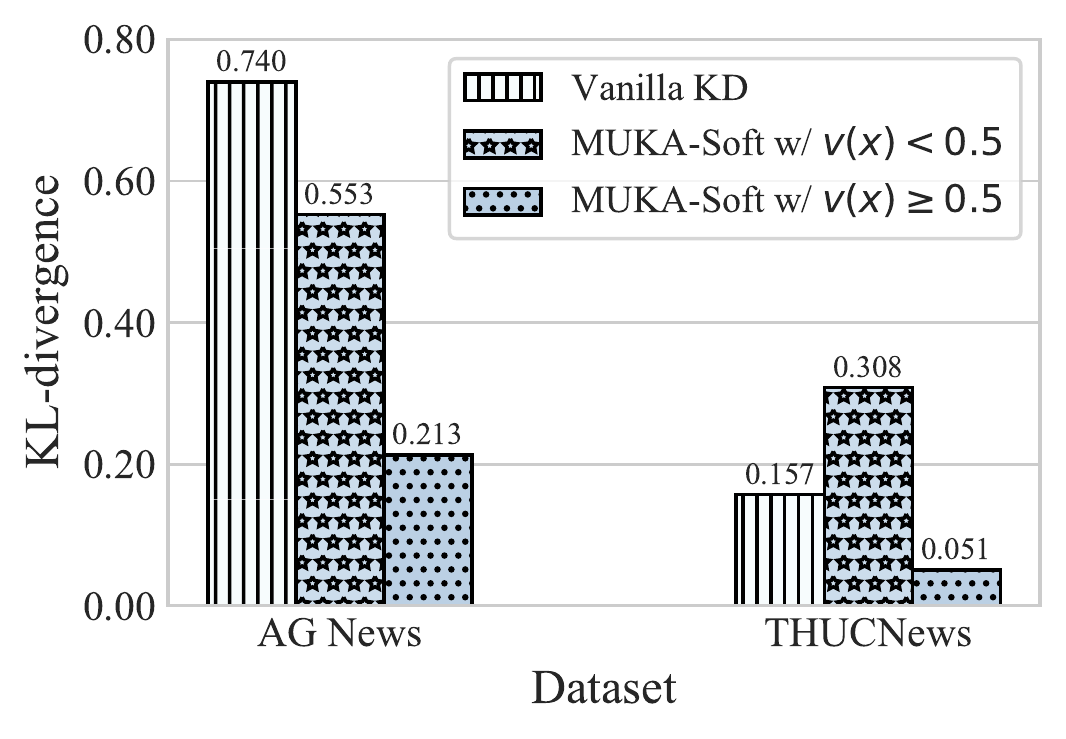}
    \caption{The KL-divergence to the golden supervision of different methods. Lower is better.}
    \label{fig:kl_analysis}
\end{figure}

\paragraph{Investigation on Supervision Quality} We probe the quality of estimated label distribution over the union label set. 
Specifically, we train a PLM with all labeled data as an oracle teacher, which thus can provide golden supervision over the union label set.
We then calculate the KL-divergence between the golden probability distribution and the approximated supervision of different combination methods. 
The lower KL-divergence indicates the combined predictions are more correct and thus can guide the training of student models better. 
We discard results of MUKA-Hard as KL-divergence is not defined for distributions with zeros, and divide MUKA-Soft into two groups according to $v(x)$, i.e. instances with $v(x) \geq 0.5$ and that with $v(x) < 0.5 $.
As shown in Figure~\ref{fig:kl_analysis},
we find that the predictions of instances with greater confidence score margin, i.e., $v(x) \geq 0.5$, are much closer to the golden distributions than the Vanilla KD.
This indicates that the estimated supervision of instances with a clearer confidence margin is of higher quality, thus paying more attention to these instances is effective for amalgamation.

\begin{table}[t!]
    \centering
    \small 
    \begin{tabular}{@{}l|c cc@{}}
    \toprule 
    \multirow{2}{*}{\textbf{Method}}     &  \textbf{3 Teachers}& \textbf{4 Teachers} & \textbf{5 Teachers}  \\
     & \{3,3,4\} & \{2,2,2,4\} & \{2,2,2,2,2\}\\ 
    \midrule
    Teacher 1 & 29.9 $\pm$ 0.00  & 20.0 $\pm$ 0.00 & 20.0 $\pm$ 0.00 \\
    Teacher 2 & 29.8 $\pm$ 0.00&  19.1 $\pm$ 0.00 & 19.1 $\pm$ 0.00\\
    Teacher 3 & 39.8 $\pm$ 0.00& 19.9 $\pm$ 0.00& 19.9 $\pm$ 0.00\\
    Teacher 4 & N / A & 39.8 $\pm$ 0.00& 20.0 $\pm$ 0.00\\
    Teacher 5& N / A & N / A & 20.0 $\pm$ 0.00\\
    Ensemble &  91.0 $\pm$ 0.00 & 74.0 $\pm$ 0.00& 80.6 $\pm$ 0.00\\
    \midrule 
    Vanilla KD  &92.5 $\pm$ 1.13 &76.5 $\pm$  0.57& 82.6 $\pm$ 1.64\\ 
     DFA &  91.1 $\pm$	1.25& 79.6 $\pm$	1.34& 82.0 $\pm$		2.04\\ 
    CFL &  93.9 $\pm$ 1.07&77.4  $\pm$ 1.34& 84.0 $\pm$ 0.24\\
    UHC & 84.4 $\pm$ 0.91& 71.6  $\pm$ 2.86 & 69.4 $\pm$ 2.73\\ 
    \midrule 
    MUKA-Hard & \textbf{94.7} $\pm$ 0.20 &93.4$^*$\hspace{-\lsuperstar}  $\pm$ 0.58& \textbf{90.5}$^*$\hspace{-\lsuperstar}  $\pm$ 0.32\\ 
    MUKA-Soft & \textbf{94.7} $\pm$ 0.14 & \textbf{93.6}$^*$\hspace{-\lsuperstar}  $\pm$ 0.30& \textbf{90.5}$^*$\hspace{-\lsuperstar} $\pm$ 1.13 \\ 
    \bottomrule
    \end{tabular}
    \caption{Results of merging multiple teacher models on the THUCNews dataset. The best results are shown in bold and $^*$ denotes the improvement over the best performing baseline is significant with $p < 0.05$. }
    \label{tab:multiple_teacher}
\end{table}

\subsection{Results in Challenging Settings}
\paragraph{KA with Multiple Teachers} As our framework is agnostic to the number of teacher models, we further conduct experiments with more teacher models. 
We conduct experiments on the THUCNews dataset as it has $10$ classes, allowing us to train up to $5$ teacher models specialized in different class sets.
The number of classes split in to $\{3, 3, 4\}$, $\{2, 2, 2, 4\}$ and $\{2, 2, 2, 2, 2\}$ for $3$, $4$ and $5$ teachers, respectively.
As shown in Table~\ref{tab:multiple_teacher}, the proposed MUKA framework generalizes well to this setting by outperforming previous baselines with a  clear margin.
Besides, we find that the previous models perform poorly under the $4$-teacher scenario.
We attribute it to that when some teacher models have more classes than others, they usually produce a larger range of logits to make the prediction more distinguishable.
Therefore, directly combing the teacher logits will lead to a biased probability distribution. 
The MUKA instead estimates the golden probability distribution according to model uncertainty scores, thus can produce better supervision even teacher models exhibit different logit scales.

\paragraph{KA with Heterogeneous Teachers}
We further consider merging knowledge from teacher models with different structures.
Specifically, we select BERT-base~($12$ layers and $768$ hidden units) and BERT-large~($24$ layers and $1024$ hidden units) as the teachers, respectively. 
The results are listed in Table~\ref{tab:heter_arch}.
We find that while a larger teacher tends to perform better, the student model performs worse on the THUCNews dataset than learning from two BERT-base teachers, indicating it is challenging to amalgamate knowledge in this setting.
Our MUKA achieves the best results on the two datasets, showing its effectiveness for heterogeneous teachers.
\begin{table}[t!]
    \centering
    \small 
    \begin{tabular}{@{}l|c  c@{}}
    \toprule 
    \multirow{2}{*}{\textbf{Method}} &  \textbf{\emph{$\text{T}_1$: BERT-base}}   & \textbf{\emph{$\text{T}_2$: BERT-large}}    \\
    &  \textbf{AG News} & \textbf{THUCNews} \\ 
    \midrule
    Teacher 1 &49.7 $\pm$ 0.00  & 48.8 $\pm$ 0.00 \\
    Teacher 2 &  47.2 $\pm$ 0.00& 49.8 $\pm$ 0.00  \\
    Ensemble & 76.6 $\pm$ 0.00 & 79.3 $\pm$ 0.00\\
    \midrule 
    Vanilla KD  &79.6  $\pm$ 0.22 &82.9 $\pm$ 1.36\\ 
    DFA & 78.2 $\pm$	0.30 & 84.7 $\pm$ 1.74  \\ 
    CFL &  75.9 $\pm$ 0.63&  81.9 $ \pm$ 1.37 \\
    UHC & 78.3 $\pm$ 	2.65& 92.3  $\pm$  1.08\\ 
    \midrule 
    MUKA-Hard &78.3 $\pm$ 	1.58 & \textbf{95.4}$^*$\hspace{-\lsuperstar} $\pm$ 0.45\\ 
    MUKA-Soft & \textbf{80.6} $\pm$  0.17& \textbf{95.4}$^*$\hspace{-\lsuperstar} $\pm$  0.29\\ 
    \bottomrule
    \end{tabular}
    \caption{Results of merging from heterogeneous teacher models with different architectures. $^*$ denotes results are statistically significant with $p < 0.05$.}
    \label{tab:heter_arch}
\end{table}
\paragraph{KA with Cross-Dataset Teachers}
Specifically, we fine-tune teacher models on different datasets separately and then train a student to perform classification over the union label set of both datasets.
The results of merging knowledge from two English datasets, AG News and Google-Snippets and even cross-lingual datasets, AG News~(in English) and THUCNews~(in Chinese) are listed in Table~\ref{tab:heter_data}.
The multilingual BERT-base are adopted for the teachers and the student in the cross-lingual setting.
Our MUKA still outperforms previous baseline models in both settings.
Interestingly, we find that the MUKA-Hard is consistently better than MUKA-Soft in this setting. We speculate the reason is that the correlations between classes of different datasets are weak, thus modeling the label relation in these disjoint groups is unnecessary.

\begin{table}[t!]
    \centering
    \small 
    \begin{tabular}{@{}l| cc  @{}}
    \toprule 
    \multirow{2}{*}{\textbf{Method}} &  \textbf{\emph{$\text{T}_1$: AG News~(en)}}   & \textbf{\emph{$\text{T}_1$: AG News~(en)}} \\
    &  \textbf{\emph{$\text{T}_2$: GS~(en)}} &  \textbf{\emph{$\text{T}_2$: THUCNews~(zh)}} \\ 
    \midrule
    Teacher 1 &26.8 $\pm$ 0.00 &  0.50 $\pm$ 0.00\\
    Teacher 2 &  63.0 $\pm$ 0.00& 54.6 $\pm$ 0.00  \\
    Ensemble & 74.3 $\pm$ 0.00& 54.5 $\pm$ 0.00 \\
    \midrule 
    Vanilla KD  &75.1  $\pm$ 0.58 &   54.6  $\pm$ 0.05\\ 
    DFA & 74.7 $\pm$	0.30  & 54.1 $\pm$ 0.20 \\ 
    CFL & 74.0 $\pm$ 0.50 & 54.3  $\pm$ 0.65\\
    UHC & 72.0 $\pm$ 0.73& 53.5  $\pm$ 0.17\\ 
    \midrule 
    MUKA-Hard & \textbf{76.3} $\pm$ 0.54&  \textbf{68.1}$^*$\hspace{-\lsuperstar}  $\pm$ 0.20\\ 
    MUKA-Soft & 75.9 $\pm$  0.47& 65.6$^*$\hspace{-\lsuperstar}   $\pm$ 0.40\\ 
    \bottomrule
    \end{tabular}
    \caption{Results of merging from teacher models with different knowledge domains and even in a cross-lingual scenario. $^*$ denotes results are statistically significant with $p < 0.05$. GS is short for Google Snippets.}
    \label{tab:heter_data}
\end{table}

\subsection{Error Analysis}
\begin{figure}[t!]
    \centering
    \includegraphics[width=1.0\linewidth]{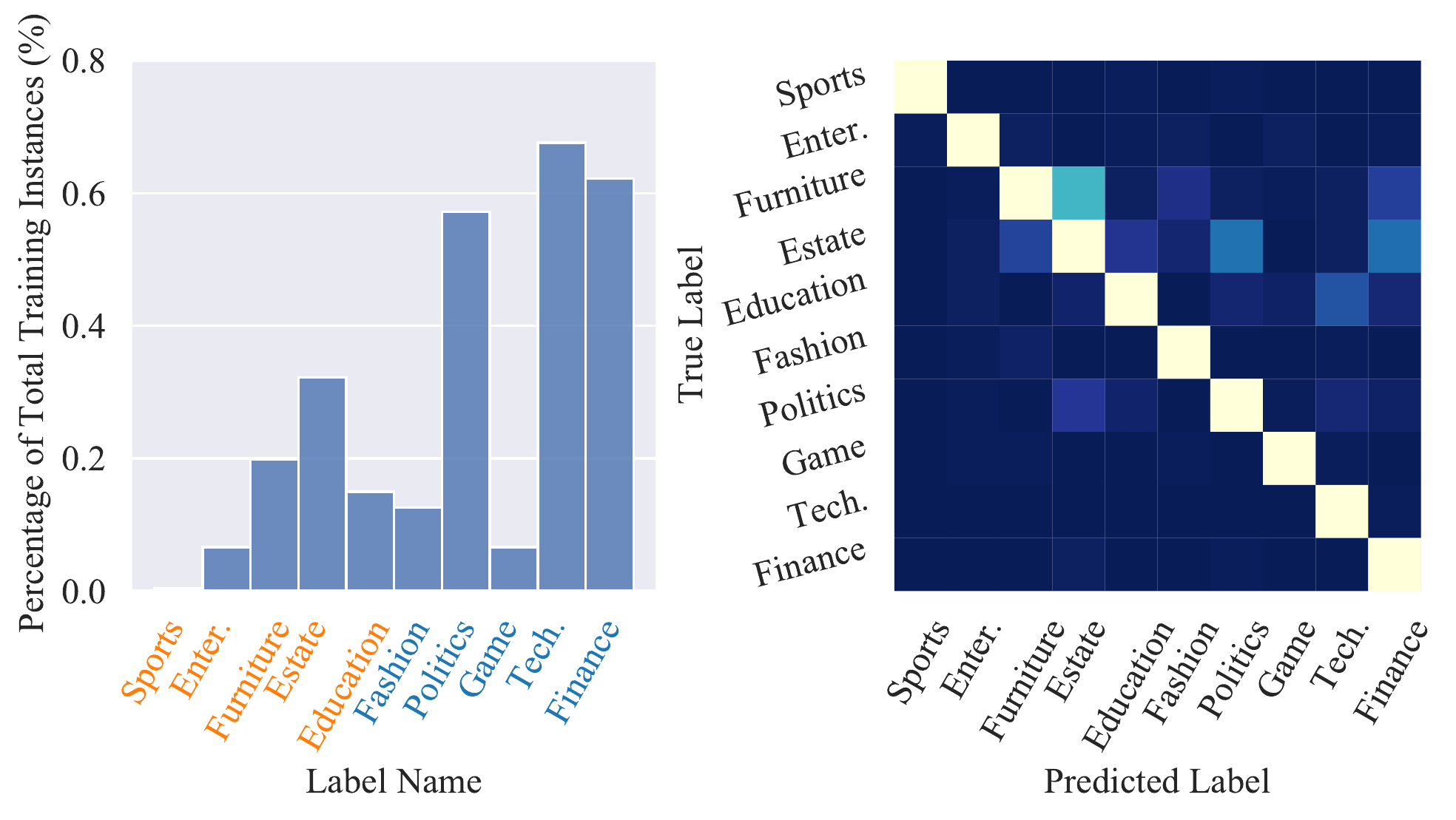}
    \caption{(Left) Label distribution of wrongly predicted instance. Labels in the same color indicate they belong to the same teacher specialty.
    (Right) The confusion matrix of an oracle model trained with all labeled data.}
    \label{fig:error_analysis}
\end{figure}
We investigate the failure cases where MUKA selects the wrong teacher model for a better understanding of MUKA.
Specifically, we probe the label distribution of instances on which MUKA assigns a higher uncertainty score to the correct teacher on THUCNews. 
As shown in the left part of Figure~\ref{fig:error_analysis}, MUKA only makes incorrect selections on a small portion of instances, and interestingly, the labels of these instances are not uniformly distributed.
To explore this, we plot the label confusion matrix of an oracle model that is fine-tuned with labeled data with all categories in the right part of Figure~\ref{fig:error_analysis}.
On the one hand, we find that there are labels that can even confuse the oracle model, e.g., instances of \emph{Estate} are tending to be classified into \emph{Politics} and \emph{Finance}, indicating that the teacher selection error of MUKA is partially due to the inner class similarity between the class labels.
On the other hand, while the oracle hardly makes wrong predictions on \emph{Tech.} and \emph{Finance} instances, the uncertainty estimation fails more frequently on these two classes than other classes. 
We leave the investigation on this phenomenon for future work.
Our instance re-weighting mechanism is effective for these instances with wrongly selected supervision source, as the the average teacher uncertainty margin $v(x)$ is $0.04$, which greatly reduces the negative impact for the student.

\section{Conclusion}
In this paper, we explore knowledge amalgamation for pre-trained language models for promoting better model reuse.
We present a principled framework MUKA, which combines the partial predictions of teachers according to the corresponding model uncertainty, and dynamically adjusts the contribution of each instance by paying more attention to instances with clearer supervision.
Experiments on text classification benchmarks demonstrate our MUKA can substantially outperform strong baselines.
Further investigations show that MUKA is generalizable for challenging settings, including merging knowledge from multiple teacher models, heterogeneous teachers, or even cross-dataset teachers.

\section{Ethical Considerations}
Our work faces several ethical challenges.
As the released PLMs may exhibit potential biases against specific groups, e.g., gender or ethnic minorities~\citep{kurita2019measuring,kennedy2020contextualizing}, these social biases can be propagated to the merged student model during the knowledge amalgamation process. 
Besides, users may collect unlabeled data from the web for conducting knowledge amalgamation, which possibly contains offensive content and thus may introducing new biases into the amalgamated model as well.

We offer possible remedies to reduce the concerns.
For the biases exhibited in the teacher models,  de-biasing techniques~\citep{zmigrod2019cda,liang2020towards,schick2020self} can be applied to eliminate the potential biases in the teachers before using the teachers model for amalgamation.
For the offensive unlabeled data collected from the internet, simple template-based or human-in-the-loop data cleaning strategies can be adopted to identify and filter potential biased data.
Except for these ad-hoc techniques, developing a bias-aware knowledge transfer framework that can de-biasing the supervision for the student model is also an interesting direction, which we are glad to explore in the future.

\bibliography{anthology,custom}
\bibliographystyle{acl_natbib}

\appendix
\section{Datasets Details}
\label{apx:dataset}
Here we provide the details of datasets used in our main paper.
The label set is first sorted according to the name and will be evenly divided into subsets according to the number of teacher models. 
Table~\ref{tab:dataset} gives the dataset statistics and the class number for two teacher models experiments.
Table~\ref{tab:class_split} gives the sorted label list on each dataset.
For teacher number that cannot evenly dividing the label sets, the final label set will include the left labels.
For example, when there are $4$ teacher models are needed on THUCNews dataset, the label set will be split into  \{\emph{Sports}, \emph{Enter.}\}, \{\emph{Furniture}, \emph{Estate}\},  \{\emph{Education}, \emph{Fashion}\} and \{\emph{Politics}, \emph{Game},\emph{Tech.}, \emph{Finance}\}.

\begin{table}[ht!]
    \centering
    \small 
    \begin{tabular}{@{}l|crrc@{}}
    \toprule
      Dataset  & {\#Class} &  {\#Train} & {\#Test} &  $\{|Y|\}$ \\
      \midrule 
      AG News  & 4  & 120,000 & 7,600 & \{2, 2\}\\ 
      5Abstracts Group & 5  & 5,300& 1,000  & \{2, 3\} \\
    Google Snippets & 8 & 10,000 & 2,200& \{4, 4\} \\ 
      THUCNews & 10  & 50,000 & 10,000&  \{5, 5\}\\
    \bottomrule
    \end{tabular}
    \caption{Statistics of datasets used in our paper. $\{|Y|\}$ denotes the number of classes each teacher model specializes.}
    \label{tab:dataset}
\end{table}

\begin{table}[tb!]
    \centering
    \small 
    \begin{tabular}{@{}l| ll@{}}
    \toprule
      Dataset   & Label Order  \\
    \midrule 
     AG News    & World, Sports,  Business, Sci/Tech  \\ 
     \midrule 
     \multirow{3}{*}{THUCNews}  & Sports, Entertainment, Furniture, \\
     & Estate, Education, Fashion, \\
     & Politics, Game, Technology, Finance\\
      \midrule 
     \multirow{4}{*}{Google Snippets} &   Business, Computers, \\ 
     & Culture-Arts-Entertainment, \\
     & Education-Science, Engineering, \\
     & Health, Politics-Society, Sports \\
     \midrule
     5Abstracts Group & Business, CSAI, Law,  Sociology, Trans\\ 
\bottomrule
    \end{tabular}
    \caption{Sorted label names of datasets. Label names of THUCNews are translated into English.}
    \label{tab:class_split}
\end{table}





\section{Hyper-parameter Search for $\tau$}
\label{apx:temp}
\begin{figure}[hbt!]
    \centering
    \includegraphics[width=0.85\linewidth]{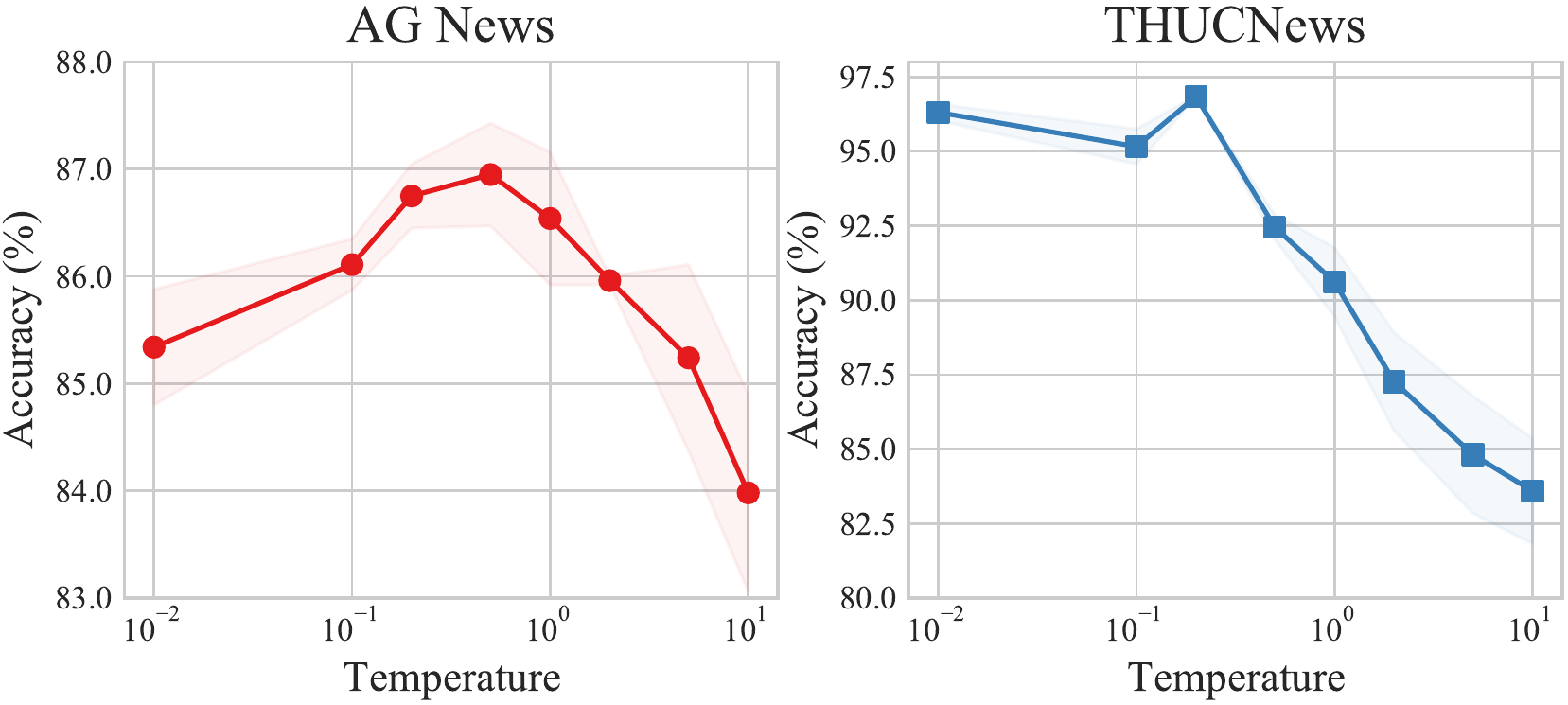}
    \caption{Varying temperature $\tau$ for MUKA-Soft. The average accuracy of three seeds are plotted with standard deviation in shade. Best viewed in color.}
    \label{fig:temperature_curve}
\end{figure}

In MUKA-Soft, $\tau$ is a hyper-parameter controlling the smoothness of teacher supervision weights. We perform a hyper-parameter search experiment for the optimal $\tau$.
We conduct experiments on AG News and THUCNews for stable results.
The values of $\tau$ are picked from $\{0.01, 0.1, 0.2, 0.5, 1.0, 2.0, 5.0, 10.0\}$, and the results with different $\tau$ are shown in Figure~\ref{fig:temperature_curve}. 
We observe that the accuracy drops significantly when $\tau$ is set to high values, where the teacher weights distribution is becoming a uniform distribution, while it reaches a peak when $\tau$ is set to a small value between $0.2$ and $0.5$.
It indicates that slightly sharpening the teacher weights distribution is helpful for the knowledge amalgamation.
Therefore, we adopt $\tau = 0.2$ in all the experiments in the main paper.

\end{document}